# High-quality Panorama Stitching based on Asymmetric Bidirectional Optical Flow

Mingyuan Meng
School of Computer Science
The University of Sydney (USYD)
Sydney, Australia
e-mail: mmen2292@uni.sydney.edu.au

Shaojun Liu
Department of Electronic and Computer Engineering
Hong Kong University of Science and Technology
Hong Kong
e-mail: eeabraham@ust.hk

*Abstract*—In this paper, we propose a panorama stitching algorithm based on asymmetric bidirectional optical flow. This algorithm expects multiple photos captured by fisheye lens cameras as input, and then, through the proposed algorithm, these photos can be merged into a high-quality 360-degree spherical panoramic image. For photos taken from a distant perspective, the parallax among them is relatively small, and the obtained panoramic image can be nearly seamless and undistorted. For photos taken from a close perspective or with a relatively large parallax, a seamless though partially distorted panoramic image can also be obtained. Besides, with the help of Graphics Processing Unit (GPU), this algorithm can complete the whole stitching process at a very fast speed: typically, it only takes less than 30s to obtain a panoramic image of 9000-by-4000 pixels, which means our panorama stitching algorithm is of high value in many real-time applications. Our code is available at https://github.com/MungoMeng/Panorama-OpticalFlow.

## I. INTRODUCTION

Panoramic images are widely used in many fields [1]. In the field of virtual reality, using panoramic images to exhibit real scenes can circumvent complex 3D modeling and rendering. In other fields such as education, tourism, and medicine, panoramic images also have their irreplaceable role. We can get wide-angle images by a wide-angle lens, but it is often difficult to get 360-degree panoramic images. Panorama stitching algorithms can stitch a set of photos with overlaps to obtain a seamless cylindrical/spherical panoramic image. With the development of image processing, panorama stitching has become an important research topic [2-4]. Most panorama stitching algorithms require the camera to rotate horizontally around a vertical rotation axis, or require multiple cameras around an axis on the same horizontal plane to capture photos at various angles at the same time. Besides, if the panorama stitching aims at a spherical panoramic image, it is also necessary to set a vertical camera to get the scene information in the vertical direction (see Figure.1).

In this paper, a panorama stitching algorithm based on asymmetric bidirectional optical flow is proposed to generate spherical panoramic images. Firstly, the distortion correction and chromaticity correction are performed on the input photos. Then, the feature points in the two adjacent photos are used to calculate a spatial transformation through which these two adjacent photos can be roughly aligned. Finally, an image-blending algorithm based on asymmetric bidirectional optical flow is used to remove the misalignments and stitching seams after the spatial transformation. The above stitching workflow is conducted iteratively until the whole set of input photos are merged into a high-quality 360-degree spherical panoramic image. Our contribution is mainly on the image-blending algorithm based on asymmetric bidirectional optical flow, and the first two steps of the above workflow are implemented with existing methods or open-source packages. We evaluated our algorithm on 20 sets of input photos and compared it with two state-of-the-art panorama stitching algorithms. The experimental results show that our algorithm outperforms the baseline algorithms on reducing stitching seams and misalignment.

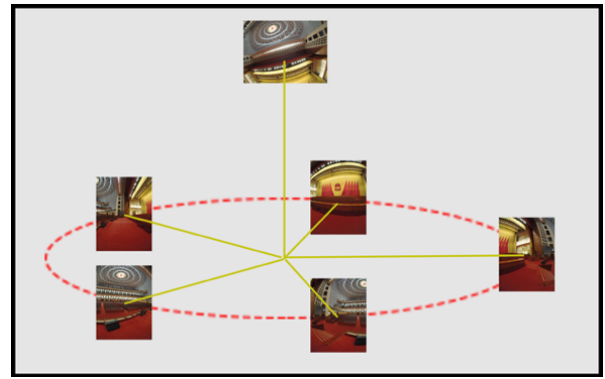

Figure 1. The relative positions of a set of input photos.

## II. RELATED WORK

Most existing photo stitching algorithms can be classified into intensity-based methods or feature-based methods [5]. Intensity-based methods are performed by establishing suitable similarity metrics to measure the similarity between two input photos. Then, by directly warping the photo, the overlaps of the two input photos are optimized to maximize their similarity. The phase correlation method proposed by Kuglin et al. [6] is a typical intensity-based stitching method. This kind of methods generally requires a relatively large overlap between adjacent photos and does not allow for local deformation, resulting in many misalignments and stitching seams. Currently, feature-based methods are relatively more popular. This kind of methods needs extract the feature points in the photos. E.g., Harris corner points [7], Speeded Up Robust Features (SURF) [8], and Scale Invariant Feature Transform (SIFT) [9]. Then, it uses the extracted feature points to find a positional relationship between the photos. Through the positional relationship, two input photos can be warped to align with each other. However, this alignment is very rough. If we directly map the warped photos on canvas and use them as the final stitching result, there will be a lot of misalignments and stitching seams. Therefore, the roughly aligned photos need to be further warped/adjusted to reduce misalignments. In this step, some widely used methods are APAP [10], SPHP [11], ANAP [12], etc. [13-14] These methods indeed improved stitching quality, but their local warp and adjustment ability are still limited, so they cannot fully solve the problem of misalignments and stitching seams.

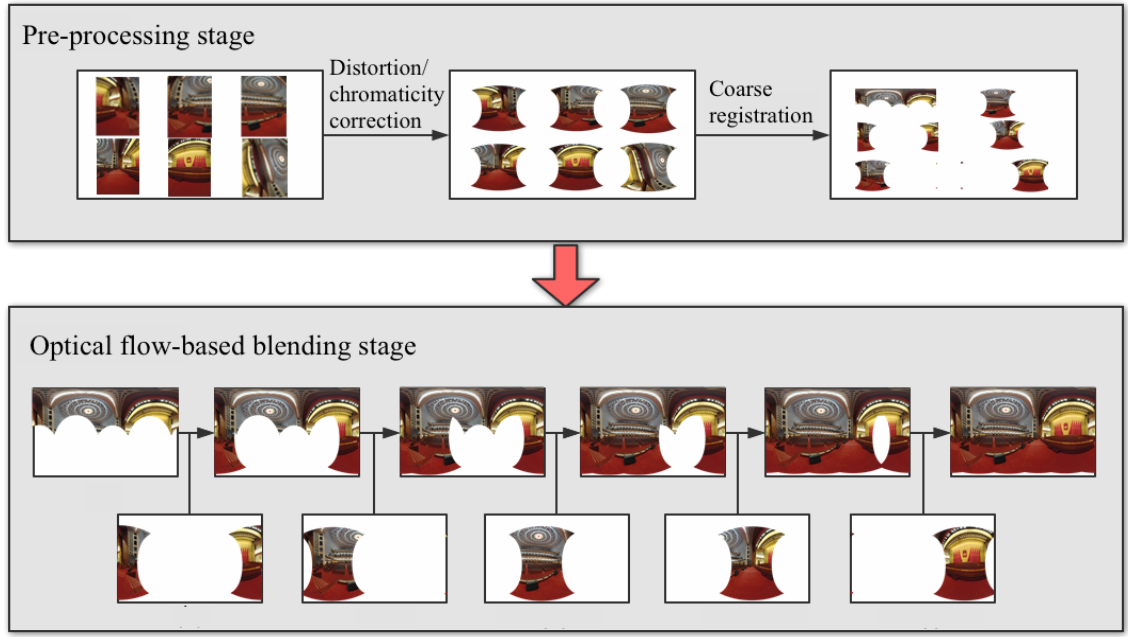

Figure 2. The workflow of our panorama stitching algorithm.

Instead of solving the above problem through local warping, we tried to use an image-blending algorithm to eliminate the misalignments and stitching seams. In this paper, we propose an image-blending algorithm based on asymmetric bidirectional optical flow. Our stitching workflow uses the feature-based stitching methods to perform a coarse registration, and then uses the information of optical flow to locally warp and adjust the photos. According to the definition of the optical flow [15], the information of optical flow can be used to warp and adjust the image at the pixel level, so this approach can further reduce misalignments and stitching seams. Compared to the previous optical flow-based stitching algorithm [16-17], ours proposed to utilize asymmetric bidirectional optical flow, which takes bidirectional information into account to reduce the biased error caused by imperfect single-directional optical flow.

III. PANORAMA STITCHING BASED ON ASYMMETRIC BIDIRECTIONAL OPTICAL FLOW

A. Workflow Overview

As is shown in Figure 2, our workflow can be divided into two stages. The first one is a pre-processing stage that can be implemented by many open-source packages or existing algorithms, including distortion correction, chromaticity correction, and coarse feature-based registration. The second one is an optical flow-based blending stage: the proposed image-blending algorithm based on asymmetric bidirectional optical flow is employed to accurately stitch each photo processed through the pre-processing stage. Since the proposed blending algorithm is designed to blend only two photos at a time, this algorithm needs to be used iteratively.

B. Pre-processing

Since the input photos are captured by fisheye cameras, we first need to correct the photos to get real images in the planar space (distortion correction). Then, since these photos facing different directions may have different exposure, we need to adjust their chromaticity (chromaticity correction). Finally, we find the rough positional relationship among them based on feature points (coarse registration). Note that each photo is output respectively along with its position information. The output of this stage will serve as the input for the next optical flow-based blending stage.

The processing at this step has been well studied and there are many existing algorithms as well as open-source software (e.g. Hugin [18], PanoTools [19], etc.) that can be adopted directly in this stage. In our experiments, we directly adopted Hugin to conduct pre-processing.

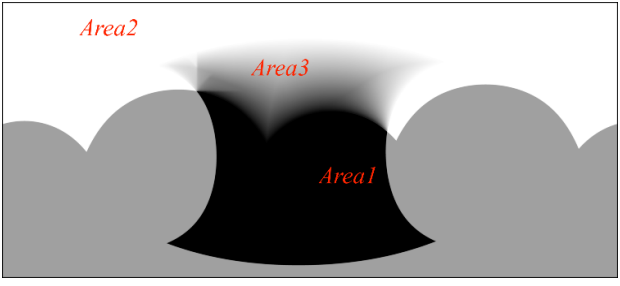

Figure 3. A visualization of global blending coefficient array *blend*

### C. Image blending based on Asymmetric Bidirectional Optical Flow

At this stage, the proposed blending algorithm based on asymmetric bidirectional optical flow is used to finely stitch the photos to eliminate the misalignments and stitching seams. The core idea of our proposed image-blending algorithm is that we try to adjust the position of each pixel in the overlapping area for a visually smooth transition where there is no visible distortion, misalignment, and stitching seams between overlapping area and other areas. Here we denote the two images to be blend as image *L* and image *R*, and the blending result as image *F*. Besides, the canvas is divided into three areas based on the positions of image *L* and *R* on the canvas: *Area1* is the area with only image *L*; *Area2* is the area with only image *R*; *Area3* is the area where image *L* and *R* overlap. We cut out the part corresponding to *Area3* in image *L* and denote it as image *OverlappedL*; similarly, the part corresponding to *Area3* in image *R* is denoted as image *OverlappedR*. Then, we do the following steps to obtain the blending result *F*.

*1) Calculate asymmetric bidirectional optical flow:* We need to calculate the optical flow of image *OverlappedR* relative to image *OverlappedL* (termed as *FlowRtoL*), as well as the optical flow of image *OverlappedL* relative to image *OverlappedR* (termed as *FlowLtoR*). Theoretically, the pair of bidirectional optical flow *FlowRtoL* and *FlowLtoR* are symmetric, but they are actually asymmetric because the algorithm to calculate optical flow itself is imperfect, so we call them asymmetric bidirectional optical flow.

Based on our review, there are many existing algorithms for calculating optical flow. E.g., Lukas-Kanade pyramid algorithm [20], Brox algorithm [21], and Farnback algorithm method [22], etc. Actually, there is not much difference in performance for these algorithms. Since the calculation for bidirectional optical flow is time-consuming, we choose the Lukas-Kanade (LK) pyramid algorithm due to its feasibility for working on GPU. The LK pyramid algorithm is a very widely-used algorithm, so its principle is not presented in detail in this paper. We can directly use OpenCV (available online at https://opencv.org/) to call a CPU version of this algorithm. In our experiments, we wrote a GPU version of LK pyramid algorithm from the scratch, and then optimize its parameters in order to obtain a more accurate bidirectional optical flow with faster execution speed. Our code is available at *https://github.com/MungoMeng/Panorama-OpticalFlow*.

```
Code 1: Blend image L and R into result image F
Input: L, R, FlowLtoR, FlowRtoL, Blend
Output: F
Defined variables: BlendL, BlendR, ColorL, ColorR,
FlowMagLR, FlowMagRL, kSoftmaxSharpness,
kFlowMagCoef, Lx, Ly, Rx, Ry, FlowL, FlowR, ExpL,
ExpR, SoftmaxL, SoftmaxR
---------------------------------------------------------------
1:  for all (i, j) in F:
2:    if (i, j) in Area1:
3:      F(i, j) = L(i, j)
4:    else if (i, j) in Area2:
5:      F(i, j) = R(i, j)
6:    else if (i, j) in Area3:
7:      BlendL = 1 – Blend(i, j)
8:      BlendR = Blend(i, j)
9:      Lx = i + FlowRtoL(i, j).x * (1 - BlendL)
10:     Ly = j + FlowRtoL(i, j).y * (1 - BlendL)
11:     ColorL = L(Lx, Ly)
12:     Rx = i + FlowLtoR(i, j).x * (1 - BlendR)
13:     Ry = j + FlowLtoR(i, j).y * (1 - BlendR)
14:     ColorR = R(Rx, Ry)
15:     FlowL = 1.0 + kFlowMagCoef * FlowMagRL
16:     ExpL = exp(kSoftmaxSharpness * BlendL * FlowL)
17:     FlowR = 1.0 + kFlowMagCoef * FlowMagLR
18:     ExpR = exp(kSoftmaxSharpness * BlendR * FlowR)
19:     SoftmaxL = ExpL/(ExpL + ExpR)
20:     SoftmaxR = ExpR/(ExpL + ExpR)
21:     F(i, j) = ColorL * SoftmaxL + ColorR * SoftmaxR
22:   else:
23:     F(i, j) = 0
```

*2) Calculate global blending coefficient:* The global blending coefficient is a concept proposed in this paper. We use it to measure the relative distance between the pixels in *Area3* and the pixels in *Area1*/*Area2*. Here we define a two-dimensional array *Blend* whose size is equal to the canvas size of image *L*/*R* to represent the blending coefficient at each pixel position. $Blend_{ij}$ represents the element of $i^{th}$ row and $j^{th}$ column of *Blend*. The values of *Blend* are defined as follows:

$$Blend_{ij} = \begin{cases} 0 & if\ (i,j) \in Area1 \\ 1 & if\ (i,j) \in Area2 \\ \frac{Lmin_{ij}}{Lmin_{ij}+Rmin_{ij}} & if\ (i,j) \in Area3 \\ 0.5 & else \end{cases} \quad (1)$$

where $Lmin_{ij}/Rmin_{ij}$ is the minimum distances from point $(i, j)$ to *Area1*/*Area2*. Figure 3 shows a visualization of array *Blend* through multiplying it by 255. The higher brightness means larger *Blend* values.

The basic idea of Blend is: the closer the position of this point is to Area1/Area2, the closer its Blend value is to 0/1.

The coefficient array *Blend* is crucial in the subsequent image-blending step. When calculating the pixel value of image *F*, the value of *Blend* will determine how each pixel is influenced by the images *L* and *R*.

*3) Blend images:* We combine the calculated asymmetric bidirectional optical flow FlowRtoL/FlowLtoR and the global blending coefficient blend to calculate the value of each pixel in image F. The basic principle is: for each pixel in image F, if it is in Area1/Area2, its value is directly taken from the pixel of image L/R in the same position; if it's in the Area3, we need to reversely map this pixel using optical flow information to find its corresponding pixel in the image L and R, and finally calculate this pixel value by a softmax function.

The pseudo-code for calculating the pixel of image *F* is shown in Code 1. The input of Code 1 is image *L/R*, optical flow *FlowLtoR/ FlowRtoL*, and coefficient array *blend*, and the output is image *F*. Besides, some extra variables need to be defined, including *BlendL*, *BlendR*, *ColorL*, *ColorR*, etc. The hyperparameter *kSoftmaxSharpness* and *kFlowMagCoef* control the sensitivity of the softmax function to variable changes.

IV. EVALUATION

In this section, we compared our proposed algorithm with two panorama stitching algorithms: the first one is widely used APAP algorithm proposed by Zaragoza et al. [10] and the other is a stitching framework based on sparse optical flow

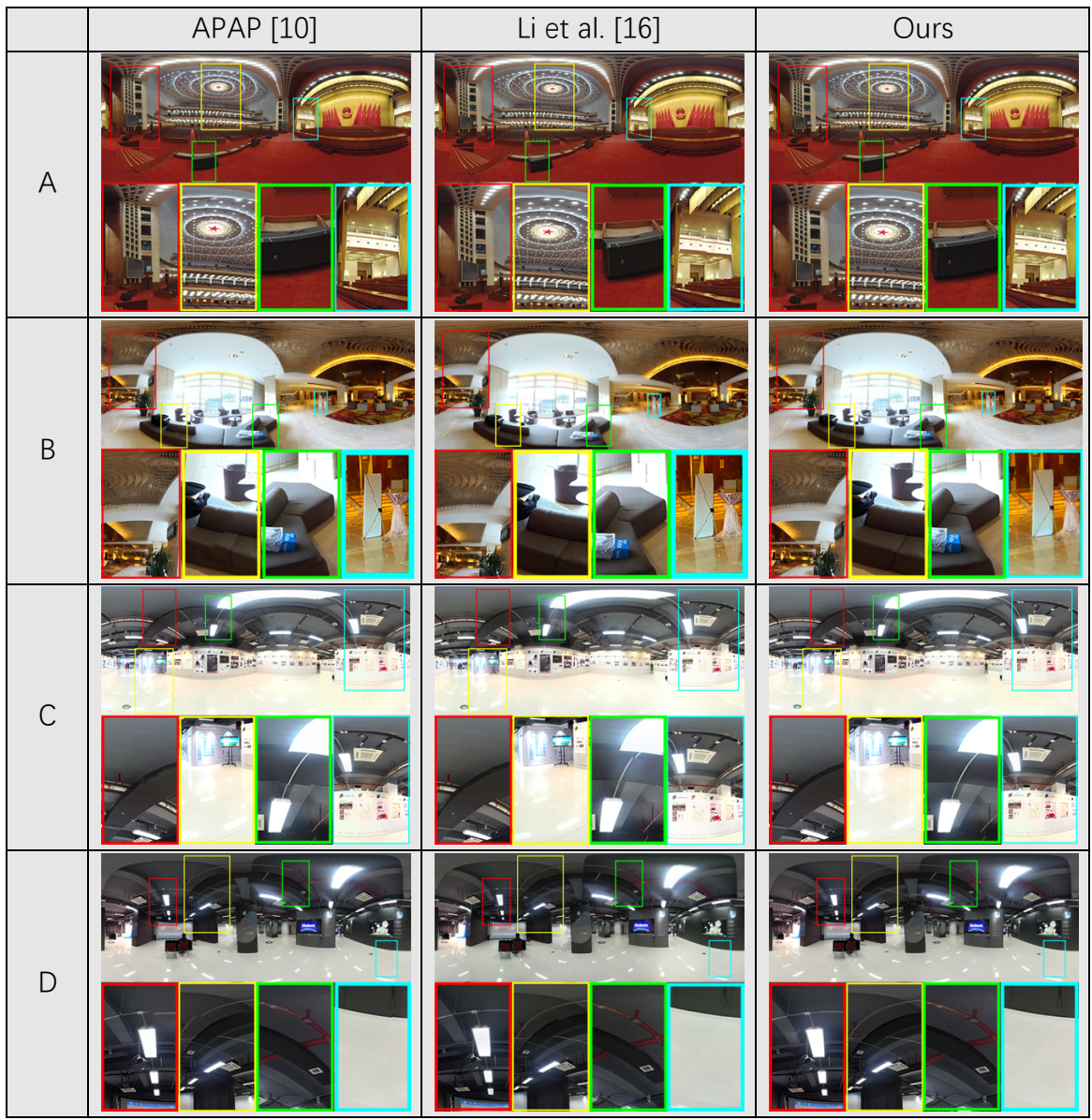

Figure 4. A visual comparison among the panoramic images produced by two baselines and our algorithm.

TABLE I. QUANTITIVE COMPARISON OF STITCHING REUSLTS

| Algorithm | A | B | C | D | average |
|---|---|---|---|---|---|
| APAP [10] | 60.5 | 33.0 | 21.5 | 35.8 | 37.7 |
| Li et al. [16] | 18.3 | 10.5 | 20.8 | 18.3 | 17.0 |
| Ours | 9.7 | 1.3 | 10.8 | 11.0 | 8.2 |

[16]. We use these three algorithms (i.e. two baselines and ours) to generate 20 panoramic images in total, and then evaluate them visually and quantitively. However, due to the space limitation of this paper, we can only exhibit 4 results.

Figure 4 shows four stitching results obtained through two baselines and our stitching algorithm. In order to perform a clearer comparison, some parts of the results are marked and enlarged in the colored box. Obviously, compared with our algorithm, the results of baseline algorithms suffer from more misalignments and stitching seams.

In addition, we also conducted a quantitative comparison of the stitching results obtained by two baselines as well as our algorithm. Each panoramic image shown in Figure 4 is in a size of 9000*4000 pixels, and there are 4 sampling positions marked by the colored box in each image. We measured the average misalignment (in pixels) in these 4 sampling positions of each panoramic image. The experimental results in TABLE I further demonstrate that our stitching algorithm outperforms the baseline algorithms on reducing misalignments.

It is also worth mentioning that our panorama stitching workflow also has an excellent performance in execution speed. In the GPU-available computing environment, it takes less than 30s to stitch a set of input photos into a panoramic image in size of 9000*4000 pixels.

## V. DISCUSSION AND CONCLUSION

In this paper, we proposed a panorama stitching algorithm that can stitch a set of input photos into a spherical panoramic image. The entire stitching workflow can be divided into a pre-processing stage and an optical flow-based fusion stage. In the pre-processing stage, we adopted existing algorithms or open-source software to pre-process the photos, including distortion/chromaticity correction and a coarse registration. In the optical flow-based blending stage, our proposed image-blending algorithm based on asymmetric bidirectional optical flow is used iteratively to finely adjust the photos at the pixel level. Our experiments in Section IV show the quality of stitching results obtained by our algorithm is obviously higher than the ones got by baseline algorithms. Besides, when GPU is available, our stitching workflow has a very fast execution speed, meaning a high value on many real-time applications.

To further improve the result of our panorama stitching workflow, we can incorporate a more advanced algorithm (e.g. [10-12]) for coarse registration in the pre-processing stage. Although these algorithms might consume much more time, it reduces the misalignments after coarse registration, and thus the quality of our final stitching result also can be further improved.

REFERENCES

[1] Wei, L. Y. U., et al. "A survey on image and video stitching." *Virtual Reality & Intelligent Hardware* 1.1 (2019): 55-83.
[2] Brown, Matthew, and David G. Lowe. "Automatic panoramic image stitching using invariant features." *International journal of computer vision* 74.1 (2007): 59-73.
[3] Gracias, Nuno, et al. "Fast image blending using watersheds and graph cuts." *Image and Vision Computing* 27.5 (2009): 597-607.
[4] Philip, Sujin, et al. "Distributed seams for gigapixel panoramas." *IEEE Transactions on Visualization and Computer Graphics* 21.3 (2014): 350-362.
[5] Szeliski, Richard. "Image alignment and stitching: A tutorial." *Foundations and Trends® in Computer Graphics and Vision* 2.1 (2006): 1-104.
[6] Kuglin, Charles D. "The phase correlation image alignment methed." *Proc. Int. Conference Cybernetics Society*. 1975.
[7] Harris, Christopher G., and Mike Stephens. "A combined corner and edge detector." *Alvey vision conference*. Vol. 15. No. 50. 1988.
[8] Bay, Herbert, Tinne Tuytelaars, and Luc Van Gool. "Surf: Speeded up robust features." *European conference on computer vision*. Springer, Berlin, Heidelberg, 2006.
[9] Lowe, David G. "Distinctive image features from scale-invariant keypoints." *International journal of computer vision* 60.2 (2004): 91-110.
[10] Zaragoza, Julio, et al. "As-projective-as-possible image stitching with moving DLT." *Proceedings of the IEEE conference on computer vision and pattern recognition*. 2013.
[11] Chang, Che-Han, Yoichi Sato, and Yung-Yu Chuang. "Shape-preserving half-projective warps for image stitching." *Proceedings of the IEEE Conference on Computer Vision and Pattern Recognition*. 2014.
[12] Lin, Chung-Ching, et al. "Adaptive as-natural-as-possible image stitching." *Proceedings of the IEEE Conference on Computer Vision and Pattern Recognition*. 2015.
[13] Vaidya, Omkar S., and Sanjay T. Gandhe. "Improvement in Image Alignment using Hybrid Warping Technique for Image Stitching." *International Journal of Engineering Research and Technology (IJERT)* 12.3 (2019): 350-356.
[14] Herrmann, Charles, et al. "Robust image stitching with multiple registrations." *Proceedings of the European Conference on Computer Vision (ECCV)*. 2018.
[15] Negahdaripour, Shahriar. "Revised definition of optical flow: Integration of radiometric and geometric cues for dynamic scene analysis." *IEEE Transactions on Pattern Analysis and Machine Intelligence* 20.9 (1998): 961-979.
[16] Li, Li, et al. "A unified framework for street-view panorama stitching." *Sensors* 17.1 (2017): 1.
[17] Labutov, Igor, Carlos Jaramillo, and Jizhong Xiao. "Fusing optical flow and stereo in a spherical depth panorama using a single-camera folded catadioptric rig." *2011 IEEE International Conference on Robotics and Automation*. IEEE, 2011.
[18] Hugin Homepage. http://hugin.sourceforge.net/ (last accessed on 01/06/2020).
[19] PanoTools Homepage. http://panotools.sourceforge.net/ (last accessed on 01/06/2020).
[20] Bruhn, Andrés, Joachim Weickert, and Christoph Schnörr. "Lucas/Kanade meets Horn/Schunck: Combining local and global optic flow methods." *International journal of computer vision* 61.3 (2005): 211-231.
[21] Brox, Thomas, and Jitendra Malik. "Large displacement optical flow: descriptor matching in variational motion estimation." *IEEE transactions on pattern analysis and machine intelligence* 33.3 (2010): 500-513.
[22] Farneback, Gunnar. "Fast and accurate motion estimation using orientation tensors and parametric motion models." *Proceedings 15th International Conference on Pattern Recognition. ICPR-2000*. Vol. 1. IEEE, 2000.